\documentclass[]{sjtu_sai}
\usepackage[toc,page,header]{appendix}
\usepackage{minitoc}
\usepackage{algorithm}
\usepackage{algpseudocode}
\usepackage{booktabs}
\usepackage{multirow}
\usepackage{siunitx}
\usepackage{makecell}
\usepackage{mathtools}
\usepackage{tabularx,array}
\usepackage[table]{xcolor}
\usepackage{graphicx}
\usepackage{wrapfig}
\usepackage{xspace}
\usepackage{amsmath}
\usepackage{amsfonts}
\usepackage{pifont}
\title{GASE: Gaussian Splatting–Based Automated System for Reconstructing Embodied-Simulation Environments}

\author[1,2,3]{Jiawei Zhang} \author[1,3]{Yiming Yan} \author[3,*]{Chao Liang}
\author[3]{Nuo Xu} \author[1,3]{Seson Sun} \author[3]{Qichen Zhang}
\author[1]{Yuhao Xu} \author[1]{Yantai Yang} \author[1]{Yingqiao Wang}
\author[2,\dagger]{Qin Jin} \author[1,\dagger]{Zhipeng Zhang}

\affiliation[1]{AutoLab, SAI, Shanghai Jiao Tong University}
\affiliation[2]{AIM3 Lab, School of Information, Renmin University of China}
\affiliation[3]{Research Lab, Anyverse Dynamics}
\contribution[*]{Project Leader}
\contribution[\dagger]{Corresponding authors}

\abstract{Training embodied agents in the real world requires skilled operators and expensive hardware. Simulation environments offer a compelling alternative by enabling large-scale, cost-effective data augmentation. Consequently, rapidly constructing high-fidelity simulation scenes with a minimal sim-to-real gap has become a critical objective in robot learning. While reconstruction-based methods provide superior visual quality, current workflows are hindered by inefficient data acquisition and subpar foreground object extraction. We thus propose GASE, a highly automated system for simulation scene construction. GASE leverages multi-view video streams from panoramic camera arrays to enable rapid environment scanning. To ensure high-quality asset generation, our pipeline introduces a camera-pose-based strategy that robustly extracts objects across frames in the 2D domain, followed by high-fidelity scene inpainting. Foreground objects and the static background are then reconstructed independently and seamlessly imported into physics simulators for policy training. Extensive experiments demonstrate that GASE outperforms existing 3D Gaussian-based methods in segmentation accuracy by over 10\% while achieving state-of-the-art inpainting quality. Furthermore, real-robot deployments across manipulation and navigation tasks maintains a performance gap of less than 10\% compared to policies trained purely on real-world data. These results confirm that GASE provides an efficient and highly effective solution for bridging the sim-to-real gap. Code will be released.}

\checkdata[Project Page]{
    \url{https://jar7visz.github.io/gase-page/}
}
\begin{document}
\maketitle

\begin{figure}[!t]
\includegraphics[width=\textwidth]{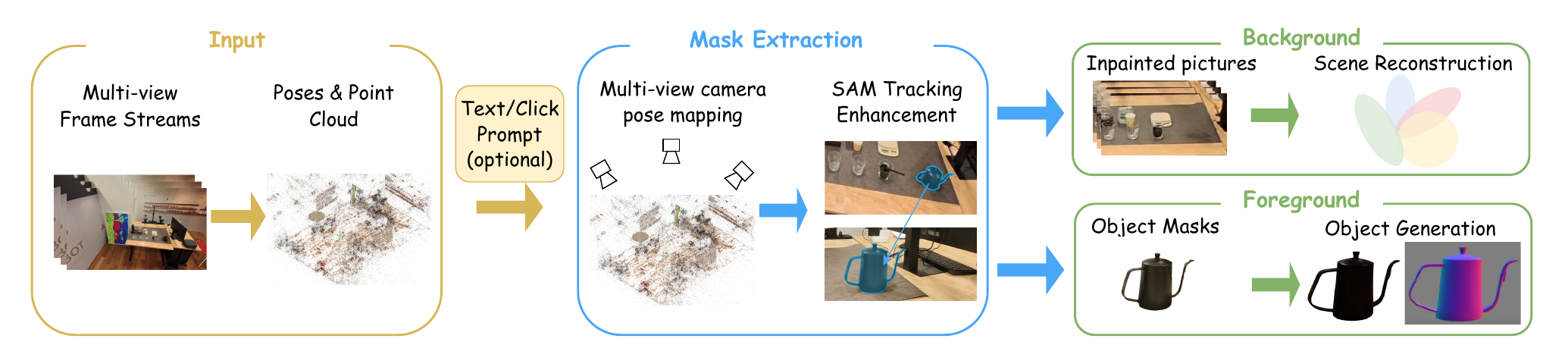}
\caption{Pipeline of our GASE system. Given scene images and camera poses and point cloud derived from them, we leverage the pose mapping relationships together with the tracking capability of SAM2 to localize target objects across all frames based on user-provided textual or click prompts, and separate them from the background. Subsequently, we respectively reconstruct the scene and the objects using 3DGS\cite{3dgs} and TRELLIS\cite{trellis}.} \label{fig1}
\end{figure}

\section{Introduction}
\label{Introduction}
As embodied artificial intelligence (AI) advances, robots are increasingly expected to execute long-horizon, multi-step tasks across diverse indoor environments. For instance, a household robot may be instructed to navigate to a bedroom, retrieve a book from a bookshelf, and deliver it to a user in the living room. However, collecting real-world demonstration data for such complex behaviors is prohibitively expensive and time-consuming. Consequently, simulation-based training has emerged as a vital paradigm. It is not only more cost-effective but also enables scalable data generation through techniques like domain randomization. This shift underscores the growing necessity of constructing high-quality simulation environments.

    To fulfill the growing demand for such training environments, existing methods primarily pursue the dual objectives of \textit{rapid scene construction} and \textit{minimizing the sim-to-real gap}. Generative approaches \cite{scenemaker,FlexWorld} typically prioritize the former by lifting a single image into a 3D scene representation via diffusion models. However, these methods frequently suffer from a substantial sim-to-real gap. Conversely, reconstruction-based methods \cite{polaris,r2r2r,re3sim,splatsim} focus on bridging this gap, which is also the primary objective of our work. These approaches typically operate as systematic pipelines. They take multi-view images and camera poses as input, reconstruct the scene using 3D Gaussian Splatting (3DGS) \cite{3dgs}, and decouple foreground objects from the static background to facilitate manipulation training. By trading off a marginal amount of construction speed, they achieve superior visual fidelity, depth accuracy, and cross-view consistency compared to generative models.

Despite these advantages, current reconstruction workflows are hindered by two critical bottlenecks regarding \textbf{data acquisition (Inefficiency)} and \textbf{object-scene decomposition (Ineffectiveness)}. First, existing systems are predominantly tailored for monocular video streams \cite{polaris,r2r2r,re3sim}, making the scanning of an expansive scene inherently labor-intensive and time-consuming. They currently lack the capability to leverage camera arrays composed of multiple panoramic cameras, which could otherwise capture entire environments rapidly. Second, these systems struggle to balance quality and efficiency when processing foreground objects. Some approaches \cite{polaris} require scanning the background and objects independently. This process becomes prohibitively cumbersome for multi-object tasks. Others \cite{r2r2r,re3sim} attempt to segment objects directly within the 3D Gaussian space. Because individual Gaussian ellipsoids often straddle object boundaries, this direct separation leads to poor extraction quality, structural holes in both the objects and the background, and subpar scene inpainting. Furthermore, systems designed purely for navigation \cite{EmbodiedSplat,planing} bypass this issue entirely by making no distinction between foreground and background, rendering them unsuitable for manipulation tasks.

To address the aforementioned challenges, we propose GASE, a highly automated simulation scene construction system based on 3D Gaussian Splatting \cite{3dgs,gsplat}. Given multiple video streams captured from a panoramic camera array in a single scan, GASE allows users to specify target objects for decomposition using intuitive text prompts and interactive clicks. To overcome the structural artifacts caused by direct segmentation in 3D space, our pipeline adopts a strategy that first extracts foreground objects and inpaints the background scene within the 2D image domain prior to any 3D reconstruction. Following this 2D processing, we leverage an object generation model to reconstruct the independent foreground assets and employ 3DGS to model the static background. These decoupled components can then be seamlessly imported into a physics simulator for embodied manipulation training. By integrating these steps, GASE establishes an efficient and highly automated workflow that successfully maintains a low sim-to-real gap.

To rigorously evaluate GASE, we construct several multi-view datasets by scanning diverse real-world environments with a panoramic camera array, encompassing a wide range of scene scales. We first assess the system's capabilities in foreground object extraction and background scene inpainting. Empirical results demonstrate that our approach outperforms existing methods operating directly on 3D Gaussians, including Gaussian Grouping \cite{GaussianGrouping}, LangSplat \cite{LangSplat}, and ObjectGS \cite{objectgs}, by a margin of over 10\% in segmentation accuracy. Furthermore, GASE achieves superior visual quality in scene inpainting tasks. Beyond visual metrics, we validate the practical utility of our reconstructed environments through real-robot experiments encompassing both manipulation and navigation tasks. Utilizing a real-to-sim-to-real paradigm (R2S2R), policies trained within our GASE-generated scenes achieve an overall success rate exceeding 60\%. Notably, this represents a performance gap of less than 10\% when compared to a baseline real-to-real pipeline \cite{pi0.5} (R2R), firmly validating that our approach effectively minimizes the sim-to-real gap.

In summary, the main contributions of this paper are as follows.
\textbf{(1)} We propose GASE, a highly automated simulation scene construction system that accepts multiple video streams from panoramic camera arrays and produces decoupled, simulation-ready assets for diverse embodied training tasks.
\textbf{(2)} We design a camera-pose-based extraction strategy that operates in the 2D domain. This method robustly identifies and segments objects across multiple frames, significantly improving segmentation quality over existing 3D Gaussian methods and enabling high-fidelity scene inpainting.
\textbf{(3)} We validate our pipeline through extensive real-robot experiments in manipulation and navigation. Our results demonstrate a minimal sim-to-real gap, with simulation-trained policies achieving performance within 10\% of those trained on real-world data.

\section{Related Work}
\subsection{Real-to-sim Environment Creation based on 3DGS}
3DGS\cite{3dgs} is a revolutionary method in the field of 3D reconstruction. Compared with earlier approaches\cite{nerf,deepsdf}, it greatly improves efficiency while preserving reconstruction quality, leading to its rapid adoption and spawning many 3DGS-based refinements\cite{gsplat,2DGS,4DGS,gssurvey}. Embodied-intelligence research has begun using 3DGS to reconstruct scenes efficiently and with high visual fidelity\cite{r2r2r,re3sim,EmbodiedSplat,planing,xsim,splatsim,realissim,twinaligner,DRAWER,lehome}. Previous work typically extracts meshes from 3DGS reconstructions, so scenes in simulators have physical collision geometry. After 3DGRUT\cite{3dgrt,3dgut} proposed a way to import neural radiance fields into Isaac Sim\cite{isaaclab,isaacsim}, building visually high-quality scenes inside the simulation became possible: researchers can import 3DGS results and the meshes derived from them to obtain simulated training environments that offer both high visual fidelity and physical collision.

\subsection{2D object segmentation, tracking and image inpainting}
\label{segmentation}
In 2D image–based tasks, 2D segmentation, tracking, and inpainting are tightly coupled. After Meta released the influential SAM\cite{sam} model, many follow-up works built on it. MobileSAM\cite{mobilesam} and similar effort focus on making SAM lightweight, and some works use SAM for video tracking\cite{deva,EmbodiedSAM,autoseg3d}. SAM’s successor, SAM2\cite{sam2}, also supports monocular video tracking with prompts in the form of points, boxes, and masks. There are also systems like Grounded SAM\cite{GroundedSAM}, which combine SAM with Grounding DINO\cite{groundingdino} to enable 2D semantic querying and segmentation. Meta’s recent SAM3\cite{sam3} further adds semantic-querying capability on top of its predecessors’ segmentation and video-tracking functions. Following segmentation, downstream inpainting methods\cite{lama,lamarefine,inpaintanything} can be applied to complete the images.

\subsection{3D Object Generation}
Recent 3D generative models are typically based on diffusion models\cite{diffusion,DeepUnsupervised} or autoregressive models\cite{openai} and support various representations such as point clouds\cite{diffpoints,pointE}, voxel grids\cite{diffvolex,diffrf,VolumeDiffusion}, triplanes\cite{diffnerf,difftri,Rodin,RodinHD}, mesh\cite{polygen,meshanything} and 3DGS\cite{gaussiancube,GVGEN}. Because we need to reconstruct segmented objects quickly and with as much realism as possible for simulator import, our task requires object-level generation. From the user’s perspective, there are methods that take a convenient single image as input\cite{sam3d}, methods that support more comprehensive multi-view input\cite{trellis}, and methods that focus on generating articulated objects\cite{omnipart,physxanything,embodiedgen,physx3d}.

\section{Method}
In this section, we describe the GASE workflow in detail. As Fig.\ref{fig1} shows, given scene images and camera poses and point cloud derived from them, we first segment the user-specified objects and perform scene inpainting. To avoid issues such as boundary ambiguity caused by the 3D Gaussian representation, we operate on 2D images. We employ SAM3\cite{sam3} to locate and track the user-specified objects. Since an object may appear intermittently across different frame streams, existing 2D tracking models struggle to recognize it as the same object, leading to extraction failure. To address this, we design a simple yet effective strategy that leverages camera poses to locate and extract objects across all frame streams. Subsequently, we use LAMA\cite{lama,lamarefine} to perform inpainting on all images, thereby completing the separation of foreground objects and the background scene. For foreground objects, we select the most suitable images and feed them into an object generation model to obtain high-quality object meshes. Since single-image input may cause the unseen angles of the object to be inconsistent with reality, we adopt TRELLIS\cite{trellis}, which supports multi-view input. For the background scene, we first perform a reconstruction using 3DGS and then extract a mesh from it, achieving high-quality visual rendering and physical collision.

\subsection{Mask Extraction and Inpainting} 
To extract a chosen object, GASE must obtain the corresponding mask for that object across all frames. Performing mask extraction independently on every frame is time-consuming and prone to inconsistency. To avoid this, GASE lets the user segment the object and pick a mask on a single frame, then maps that selection to all frames using our localization strategy. 
% GASE supports two image-selection modes: (1) provide the image ID directly; or (2) visualize the point cloud and interactively box-select points—GASE maps the selected point set ${P}$ to all images ${F}$ and returns several candidate images ranked by coverage to allow users unfamiliar with the dataset to choose quickly. Details of candidate images ranking are in Appendix \ref{appendix a}.

\paragraph{Localization Strategy}
We propose a novel and concise strategy to solve the problem that 2D tracking models cannot reliably propagate identities across views in multiple frame streams. First, we use SAM3 to extract all masks from a selected image; the user may optionally provide a text prompt. For each frame $f$, we first transform each point $\mathbf{P}j$ into camera coordinates using extrinsics $(\mathbf{R}f,\mathbf{t}f)$, obtaining $(x_{j,f}^{c},y_{j,f}^{c},z_{j,f}^{c})$, and then project it to image coordinates $(u_{j,f},v_{j,f})$ using intrinsics $(f_x^f,f_y^f,c_x^f,c_y^f)$ as:
\begin{equation}
\begin{bmatrix}
x_{j,f}^{c}\\
y_{j,f}^{c}\\
z_{j,f}^{c}
\end{bmatrix}
=
\mathbf{R}_f\mathbf{P}_j+\mathbf{t}_f,\qquad
u_{j,f}=f_x^f\frac{x_{j,f}^{c}}{z_{j,f}^{c}}+c_x^f,\quad
v_{j,f}=f_y^f\frac{y_{j,f}^{c}}{z_{j,f}^{c}}+c_y^f .
\end{equation}
Because direct projection ignores inter-object occlusion, we filter projections using depth information from the point cloud. To handle occlusion, let $D_f(u,v)$ denote the depth map at pixel $(u,v)$ in frame $f$,
and let $\tau$ be a depth tolerance. A projected point is kept if
\begin{equation}
z_{j,f}^{c}\le D_f(u_{j,f},v_{j,f})+\tau .
\end{equation}
For each image, we set a threshold $\tau$: if the number of points projected onto that image exceeds $\tau$, we select the smallest mask that contains the largest number of projected points as the result mask $\mathbf{M_{res}}$ for that image.

\paragraph{Tracking and Inpainting}
During scene scanning, the same object may appear at different scales or only partially in different images, so the above procedure can still miss some masks. Since we know the start and end frames for each sequence, we further enhance the results using SAM2’s video propagation. For each frame stream, we take the first image that already contains a result mask $\mathbf{M_{res}^i}$  and propagate that mask with SAM2, marking all images reached by propagation. We then move on to the next image that contains $\mathbf{M_{res}^i}$  but has not been marked, and repeat the process. Subsequently, we remove all resulting masks and perform high-resolution inpainting using LAMA\cite{lama}.

\subsection{Scene Reconstruction}
\paragraph{Reconstruction based on 3DGS}
After obtaining the inpainted scene images, we reconstruct the scene using 3DGS\cite{3dgs}. Because the objects have been removed, we first use the mask–point-cloud mapping to delete the 3D points in the point cloud that correspond to the removed object. Next, we input the scene images together with the COLMAP-format data (camera parameters and the 3D point cloud) into gsplat\cite{gsplat} for reconstruction. We represent the scene as a set of $K$ 3D Gaussians:
\begin{equation}
\mathcal{G}
=
\left\{
\left(
\boldsymbol{\mu}_k,\,
\mathbf{q}_k,\,
\mathbf{s}_k,\,
o_k,\,
\mathbf{c}^{\mathrm{SH}}_k
\right)
\right\}_{k=1}^{K}.
\end{equation}

Here, $\boldsymbol{\mu}_k\in \mathbb{R}^3$ is the Gaussian center, $\mathbf{q}_k\in\mathbb{R}^4$ is the quaternion, $\mathbf{s}_k\in\mathbb{R}^3$ is the log-scale parameter, $o_k$ is the logit opacity parameter, and $\mathbf{c}^{\mathrm{SH}}_k$ denotes SH color coefficients (split as $\texttt{sh0}$ and $\texttt{shN}$ in implementation).  
The positive scale and opacity used in the rendering are
\begin{equation}
\tilde{\mathbf{s}}_k=\exp(\mathbf{s}_k),\qquad
\alpha_k=\sigma(o_k),
\end{equation}
with $\sigma(\cdot)$ the sigmoid function.
To enable subsequent mesh extraction, we render depth maps $\hat{D}$ after reconstruction is complete. Finally, we convert the Gaussian point cloud into a format importable into Isaac Sim\cite{isaaclab,isaacsim} using 3DGRUT\cite{3dgrt,3dgut}.

\paragraph{Mesh Extraction}
To give a background physical collision geometry for the simulation, we extract a mesh from the Gaussians. First, we integrate the RGB-D information from each frame into a TSDF voxel grid according to the camera poses, then extract a point cloud and normals from the TSDF and perform a Poisson surface reconstruction\cite{Poisson,ScreenedPoisson} to minimize surface holes. For small scenes that are less prone to holes but require higher detail, we offer an alternative: extract a triangle mesh from the TSDF zero level set, then apply post-processing such as hole filling, smoothing, and connected-component cleanup. This yields a high-fidelity 3DGS background together with a mesh that provides physical collision properties.

\subsection{Object Generation}
We input the extracted object masks to TRELLIS\cite{trellis} for object generation. To save resources while preserving object detail as much as possible, we select the most suitable subset of masks as input. We design a weighting scheme, sort all masks by weight in descending order, and then greedily select masks. To ensure comprehensive coverage of the object, we require that any two consecutively selected masks be at least $d$ frames apart. Weighting scheme details are in App. \ref{appendix c}.

\section{Experiments}
\label{sec:result}
\subsection{Setups}
We used several panoramic cameras to create a dataset composed of multiple frame streams, spanning a range of scene scales. It includes 4 tabletop scenes and 5 indoor scenes ranging in size from approximately 10 to 100 square meters.
 Manuplation tasks are tested on the piper robot arm\cite{piper_arm} and navigation tasks are tested on the Anyverse Kitt 15 robot. All experiments are tested on a single NVIDIA RTX A800, and the simulation environment used Isaac Sim 5.1\cite{isaacsim}. To validate the reliability of 2D-based mask extraction and inpainting, we conducted experiments comparing them with 3D-based methods. All methods were trained for 30k iterations with 3DGS.

\subsection{Extraction and Inpainting}

\begin{table}\centering\small
\caption{mIoU and mBIoU results on the LERF dataset(\%).}\label{tab1}
\begin{tabular}{lcccccc}
\toprule
  & \multicolumn{2}{c}{figurines} & \multicolumn{2}{c}{ramen} & \multicolumn{2}{c}{teatime} \\
\cmidrule(lr){2-3}\cmidrule(lr){4-5}\cmidrule(lr){6-7}
  & mIoU$\uparrow$ & mBIoU$\uparrow$ & mIoU$\uparrow$ & mBIoU$\uparrow$ & mIoU$\uparrow$ & mBIoU$\uparrow$ \\
\midrule
LangSplat\cite{LangSplat} & 54.70 &50.20  &86.32  &75.79  & 64.06 &55.79  \\
Gaussian Grouping\cite{GaussianGrouping} &77.76  &73.86  &61.50  &58.71  &89.43  &86.30  \\
ObjectGS\cite{objectgs} &78.41  &77.73  &84.89  &82.35  & 68.94 &66.43\\
Ours & \textbf{93.32} & \textbf{89.59} & \textbf{95.63} & \textbf{94.34} & \textbf{95.91} & \textbf{92.61} \\
\bottomrule
\end{tabular}
\end{table}

\paragraph{Extraction} For extraction, we conducted comparative experiments on the LERF dataset against 3DGS-based methods to verify that the 2D-mask-based approach performs semantic segmentation more accurately and with clearer boundaries, as shown in Tab. \ref{tab1}. Same as Gaussian Grouping\cite{GaussianGrouping}, for each category ID, we first average IoU (and separately BIoU) over all evaluated images for that category, then report the macro average over categories. Experiments show that GASE’s 2D-based segmentation method achieves over 10\% higher performance than LangSplat\cite{LangSplat}, Gaussian Grouping\cite{GaussianGrouping}, and ObjectGS\cite{objectgs}.

\begin{figure}[htbp]
\centering
\includegraphics[width=\textwidth]{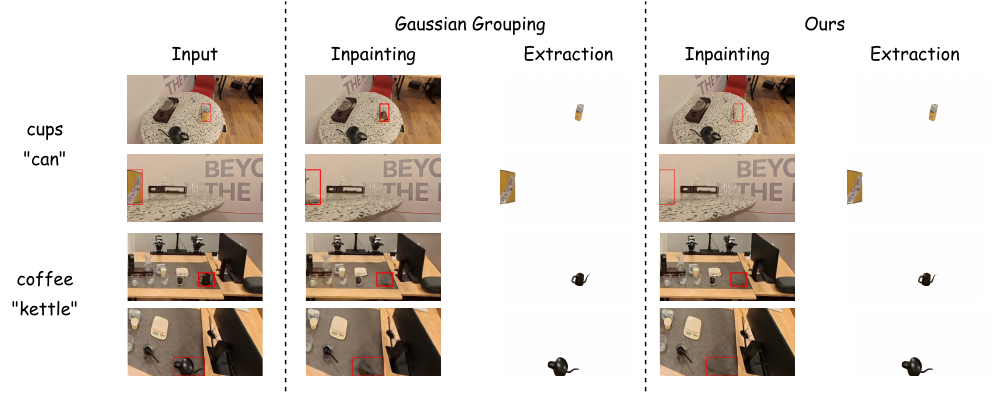}
\caption{Results of extraction and inpainting.} \label{fig2}
\end{figure}

\paragraph{Inpainting} For inpainting, we conducted experiments in the sundries scenario of our dataset. The scene images after LaMa inpainting were used as input for 3DGS training, and the resulting inference images were compared with those produced by other methods. The results are shown in the Fig. \ref{fig2} and Tab. \ref{tab2}. Experimental results show that, compared with methods that perform inpainting directly within Gaussian representations such as Gscream\cite{gscream}, Gaussian Grouping\cite{GaussianGrouping}, and 3DGIC\cite{3dgic}, our approach, which conducts inpainting on 2D images prior to scene reconstruction, achieves state-of-the-art performance in both visual quality and consistency.

\begin{table}\centering\small
\caption{Rendering and inpainting results in our sundries scene. Scores begin with m- are only calculated within the ground truth inpainting masks.}\label{tab2}
\begin{tabular}{lcccccc}
\toprule
  &  PSNR$\uparrow$  & SSIM$\uparrow$ & LPIPS$\downarrow$ & m-LPIPS$\downarrow$& FID$\downarrow$ & m-FID$\downarrow$\\
\midrule
GScream\cite{gscream} & 14.08  &0.80  &0.49  &0.14  &412.67 &434.34  \\
Gaussian Grouping\cite{GaussianGrouping}&26.22   &0.93  &\textbf{0.15}  &0.14  &180.97 &332.86  \\
3DGIC\cite{3dgic} &24.91   &0.91  &0.16  &0.14  &168.68 &330.23  \\
Ours&\textbf{27.09}   &\textbf{0.94}  &\textbf{0.15}  &\textbf{0.13}  &\textbf{166.14} &\textbf{313.80}  \\
\bottomrule
\end{tabular}
\end{table}

\subsection{Sim to Real}
To demonstrate that GASE exhibits a sufficiently small sim-to-real gap in practical use, we designed manipulation and navigation experiments both in simulation and on real hardware. 

\begin{table}\centering\small
\caption{Success rate(\%) of manipulation tasks using GASE in sundries scene. We collected 50 samples in both simulation and real world for training, and performed evaluation on the real robot. 
}\label{tab3}
\begin{tabular}{lccc}
\toprule
  &sim2sim & sim2real & real2real \\
\midrule
grasp cola can&0.67 & 0.73 &  0.73 \\
put the bottle into the box &0.60 &  0.60&  0.67 \\
push the cup over &0.67 & 0.67 &   0.67 \\
\bottomrule
\end{tabular}
\end{table}

\begin{figure}
\includegraphics[width=\textwidth]{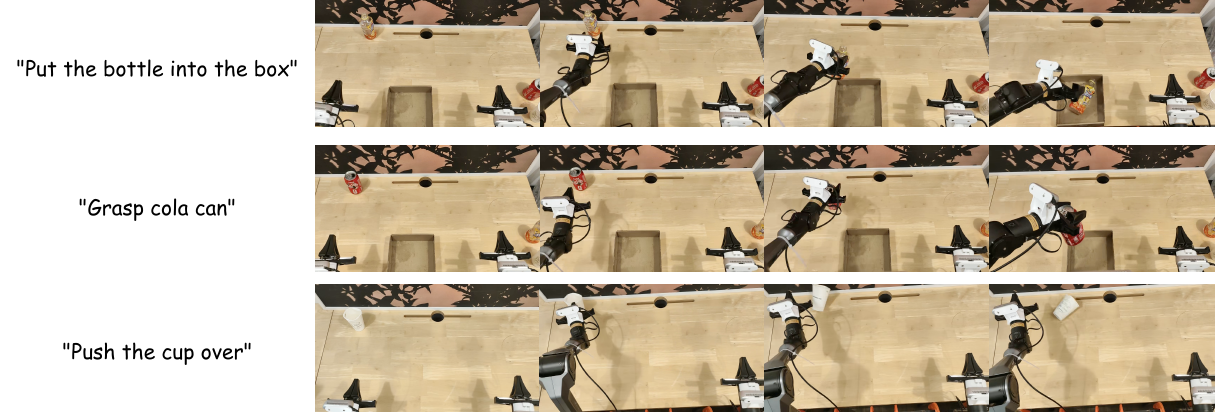}
\caption{Experiment on manipulation tasks.} \label{manip1}
\end{figure}

\paragraph{Manipulation} For the manipulation task, we deployed Pi 0.5\cite{pi0.5} on the Piper robotic arm and tested it in the ``sundries'' scenario of our dataset. As shown in Tab.\ref{tab3} and Fig.\ref{manip1}, we designed three tasks that involve objects of different materials: (1) \textit{Grasp cola can}. Grasp a metal cola can and move it to the area in front of the overhead camera above the cardboard box. (2) \textit{Put the bottle into the box}. Grasp a plastic bottle and place it in the  box. (3) \textit{Push the cup over}. Pushing over a deformable paper cup. For each task, 50 episodes of data samples were collected in both simulation and real-world settings for training and evaluation. We tested three paradigms, namely sim-to-sim, sim-to-real, and real-to-real. The results show that the sim-to-real success rate exceeds 60\%, and the performance gap relative to real-to-real is within 10\%, demonstrating that GASE exhibits a small sim-to-real gap. 

\begin{figure}
\includegraphics[width=\textwidth]{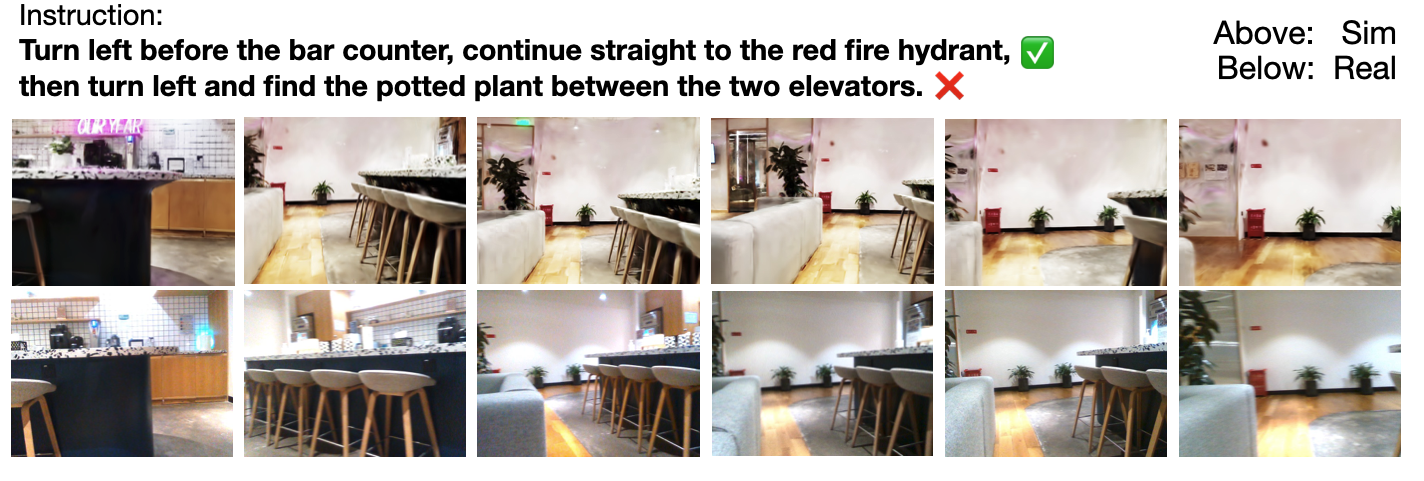}
\caption{Experiment on navigation tasks.} \label{nav1}
\end{figure}

\paragraph{Navigation} For the navigation task, we deployed Uni-NaVid\cite{uninavid} on the Kitt 15 robot and tested in the ``bar'' scenario of our dataset. As shown in Tab.\ref{tab5}, we designed three tasks: (1) \textit{Find the microwave}. Start from the elevator area and locate the microwave. (2) \textit{Find a red fire extinguisher box}. Start from the right side of the bar and locate the fire extinguisher box even though it is not visible in the initial frame. (3) \textit{Straight to the wall, then turn left}. Start from the right side of the bar, move toward the wall, and then turn left. We evaluated the zero-shot performance of the model in real-world settings, as well as its performance in both simulation and real-world settings after training in simulated data. For each task, 100 samples were collected. As shown by the results, training with simulated data significantly improves the success rate. Fig.\ref{nav1} also demonstrates that the GASE system exhibits a minimal sim-to-real gap: for long-range tasks, the algorithm shows consistent numbers of successful and failed trials in both simulation and real-world settings. 

\begin{table}\centering\small
\caption{Success rate(\%) of navigation tasks using GASE in wework scene. We evaluated the same model in the same scenario by comparing its success rates before and after training in the simulation environment. Because physical collisions in simulation are more stringent, the sim-to-sim success rate is lower than the sim-to-real success rate.}\label{tab5}
\begin{tabular}{lccc}
\toprule
  & zero-shot in real & sim2sim & sim2real  \\
\midrule
find the microwave & 75  &83.33 &  100  \\
find a red fire extinguisher box &0 &66.67   & 100  \\
straight to the wall, then turn left & 100 &50  & 100   \\
\bottomrule
\end{tabular}
\end{table}

\subsection{Comparison with other systems}
Tab.\ref{tab7} presents a comparison of GASE with other related systems. The comparative results indicate that our system provides greater convenience, a higher level of automation, and a broader applicability across scene scales.

\begin{table}\centering\small
\caption{Comparison between GASE and other systems. R2R2R denotes real2render2real\cite{r2r2r}. PLANNING\cite{planing} does not consider foreground objects. The scene-scale categories follow the same classification as in our dataset: small refers to tabletop scenes, medium refers to small indoor scenes, and large refers to large indoor scenes.}\label{tab7}
\begin{tabular}{lccccc}
\toprule
  & PolaRiS & Re3Sim  & PLANING& R2R2R & GASE(Ours) \\
\midrule
Gaussian representation& \ding{51}  & \ding{51}  & \ding{51} & \ding{51} & \ding{51}  \\
No additional object scanning & \ding{55}  & \ding{51}  & \ding{55} & \ding{51} & \ding{51} \\
Supports multiple frame streams input & \ding{55}  & \ding{55} & \ding{55} & \ding{55} & \ding{51} \\
\bottomrule
\end{tabular}
\end{table}

\subsection{Ablation}
In this section, we present ablation studies designed to validate the necessity and feasibility of key components. To validate our cross-view propagation strategy, we measure the difference in the proportion of frames successfully extracted (relative to the total number of frames that should be extracted) under three conditions: SAM3 tracking only, COLMAP mapping only, and COLMAP+SAM3. Additionally, to validate the efficiency of our strategy, we compared it's performance with that of performing open-vocabulary semantic query on every frame using SAM3. The results are shown in Tab.\ref{tab8}.  

\begin{table}\centering
\caption{Ablation on mask extraction and segmentation(\%). We use the percentage of successfully extracted frames relative to the total number of frames that should be extracted as the metric, and test on datasets at three scales (large, medium, and small). OOM means CUDA out of memory.}\label{tab8}
\begin{tabular}{lcccc}
\toprule
scene  & SAM3 on each frame& SAM3 tracking& Colmap  & Colmap+SAM3 \\
\midrule
videogame &59.36  &77.43 &68.24  &\textbf{79.96} \\
office   &64.67 &OOM & 74.83 &\textbf{82.43} \\
bar   & 49.80 &OOM &52.37   &\textbf{67.49} \\
\bottomrule
\end{tabular}
\end{table}

\section{Conclusion}
\label{sec:conclusion}
We propose GASE, a novel 3DGS-based system for constructing embodied simulation environments. It employs a concise and effective strategy to process multiple frame streams, separating objects from the scene and reconstructing both the scene and objects for embodied simulation training. We also used panoramic cameras to build a dataset composed of multiple frame streams  and conducted experiments on it to demonstrate GASE’s high level of automation, broad applicability, and strong performance. 

\section{Limitation}
\label{limitation}
Our system assumes that point clouds and camera poses are derived from scene images using purely vision-based methods, and that foreground objects and background images are separated in the 2D domain. However, when dealing with scene images composed of multiple frame sequences, purely vision-based approaches are time-consuming, produce relatively sparse point clouds, and are prone to losing camera viewpoints. In addition, separation methods operating in the 2D domain lack real-time capability. Feed-forward reconstruction and semantic segmentation within 3DGS could potentially address these limitations; however, their current accuracy is still insufficient for practical deployment. This remains an important challenge for future research. More details and experiments can be found in App.\ref{app:limitation} and App.\ref{appendix d}.

\section{Acknowledgments}
This work was supported in part by the Natural Science Foundation of China.

\clearpage
\nocite{*}
\bibliographystyle{plainnat}
\bibliography{main}

@Article{Gauss1857,
  Title                    = {Theory of the motion of the heavenly bodies moving about the sun in conic sections},
  Author                   = {Carl Friedrich Gauss and Charles Henry Davis},
  Journal                  = {Gauss's Theoria Motus},
  Year                     = {1857},
  Number                   = {1},
  Pages                    = {5--23},
  Volume                   = {76}
}

@Book{Lagrange1788,
  title		= {M{\'e}canique Analytique},
  author	= {Joseph-Louis Lagrange},
  publisher	= {Desaint, Paris},
  year		= {1788}
}

@article{ RT1,  
 title={RT-1: Robotics Transformer for Real-World Control at Scale}, 
 author={Brohan, Anthony and Brown, Noah and Carbajal, Justice and Chebotar, Yevgen and Dabis, Joseph and Finn, Chelsea and Gopalakrishnan, Keerthana and Hausman, Karol and Herzog, Alex and Hsu, Jasmine and Ibarz, Julian and Ichter, Brian and Irpan, Alex and Jackson, Tomas and Jesmonth, Sally and Joshi, NikhilJ and Julian, Ryan and Kalashnikov, Dmitry and Kuang, Yuheng and Leal, Isabel and Lee, Kuang-Huei and Levine, Sergey and Lu, Yao and Malla, Utsav and Manjunath, Deeksha and Mordatch, Igor and Nachum, Ofir and Parada, Carolina and Peralta, Jodilyn and Perez, Emily and Pertsch, Karl and Quiambao, Jornell and Rao, Kanishka and Ryoo, Michael and Salazar, Grecia and Sanketi, Pannag and Sayed, Kevin and Singh, Jaspiar and Sontakke, Sumedh and Stone, Austin and Tan, Clayton and Tran, Huong and Vanhoucke, Vincent and Vega, Steve and Vuong, Quan and Xia, Fei and Xiao, Ted and Xu, Peng and Xu, Sichun and Yu, Tianhe and Zitkovich, Brianna}, 
 year={2022}, 
 month={Dec}, 
 language={en-US} 
 }

@misc{pi0,
      title={$\pi_0$: A Vision-Language-Action Flow Model for General Robot Control}, 
      author={Kevin Black and Noah Brown and Danny Driess and Adnan Esmail and Michael Equi and Chelsea Finn and Niccolo Fusai and Lachy Groom and Karol Hausman and Brian Ichter and Szymon Jakubczak and Tim Jones and Liyiming Ke and Sergey Levine and Adrian Li-Bell and Mohith Mothukuri and Suraj Nair and Karl Pertsch and Lucy Xiaoyang Shi and James Tanner and Quan Vuong and Anna Walling and Haohuan Wang and Ury Zhilinsky},
      year={2026},
      eprint={2410.24164},
      archivePrefix={arXiv},
      primaryClass={cs.LG},
      url={https://arxiv.org/abs/2410.24164}, 
}

@misc{pi0fast,
      title={FAST: Efficient Action Tokenization for Vision-Language-Action Models}, 
      author={Karl Pertsch and Kyle Stachowicz and Brian Ichter and Danny Driess and Suraj Nair and Quan Vuong and Oier Mees and Chelsea Finn and Sergey Levine},
      year={2025},
      eprint={2501.09747},
      archivePrefix={arXiv},
      primaryClass={cs.RO},
      url={https://arxiv.org/abs/2501.09747}, 
}

@misc{pi0.5,
      title={$\pi_{0.5}$: a Vision-Language-Action Model with Open-World Generalization}, 
      author={Physical Intelligence and Kevin Black and Noah Brown and James Darpinian and Karan Dhabalia and Danny Driess and Adnan Esmail and Michael Equi and Chelsea Finn and Niccolo Fusai and Manuel Y. Galliker and Dibya Ghosh and Lachy Groom and Karol Hausman and Brian Ichter and Szymon Jakubczak and Tim Jones and Liyiming Ke and Devin LeBlanc and Sergey Levine and Adrian Li-Bell and Mohith Mothukuri and Suraj Nair and Karl Pertsch and Allen Z. Ren and Lucy Xiaoyang Shi and Laura Smith and Jost Tobias Springenberg and Kyle Stachowicz and James Tanner and Quan Vuong and Homer Walke and Anna Walling and Haohuan Wang and Lili Yu and Ury Zhilinsky},
      year={2025},
      eprint={2504.16054},
      archivePrefix={arXiv},
      primaryClass={cs.LG},
      url={https://arxiv.org/abs/2504.16054}, 
}

@misc{pi0.6,
      title={$\pi^{*}_{0.6}$: a VLA That Learns From Experience}, 
      author={Physical Intelligence and Ali Amin and Raichelle Aniceto and Ashwin Balakrishna and Kevin Black and Ken Conley and Grace Connors and James Darpinian and Karan Dhabalia and Jared DiCarlo and Danny Driess and Michael Equi and Adnan Esmail and Yunhao Fang and Chelsea Finn and Catherine Glossop and Thomas Godden and Ivan Goryachev and Lachy Groom and Hunter Hancock and Karol Hausman and Gashon Hussein and Brian Ichter and Szymon Jakubczak and Rowan Jen and Tim Jones and Ben Katz and Liyiming Ke and Chandra Kuchi and Marinda Lamb and Devin LeBlanc and Sergey Levine and Adrian Li-Bell and Yao Lu and Vishnu Mano and Mohith Mothukuri and Suraj Nair and Karl Pertsch and Allen Z. Ren and Charvi Sharma and Lucy Xiaoyang Shi and Laura Smith and Jost Tobias Springenberg and Kyle Stachowicz and Will Stoeckle and Alex Swerdlow and James Tanner and Marcel Torne and Quan Vuong and Anna Walling and Haohuan Wang and Blake Williams and Sukwon Yoo and Lili Yu and Ury Zhilinsky and Zhiyuan Zhou},
      year={2025},
      eprint={2511.14759},
      archivePrefix={arXiv},
      primaryClass={cs.LG},
      url={https://arxiv.org/abs/2511.14759}, 
}

@misc{aloha2,
      title={ALOHA 2: An Enhanced Low-Cost Hardware for Bimanual Teleoperation}, 
      author={ALOHA 2 Team and Jorge Aldaco and Travis Armstrong and Robert Baruch and Jeff Bingham and Sanky Chan and Kenneth Draper and Debidatta Dwibedi and Chelsea Finn and Pete Florence and Spencer Goodrich and Wayne Gramlich and Torr Hage and Alexander Herzog and Jonathan Hoech and Thinh Nguyen and Ian Storz and Baruch Tabanpour and Leila Takayama and Jonathan Tompson and Ayzaan Wahid and Ted Wahrburg and Sichun Xu and Sergey Yaroshenko and Kevin Zakka and Tony Z. Zhao},
      year={2024},
      eprint={2405.02292},
      archivePrefix={arXiv},
      primaryClass={cs.RO},
      url={https://arxiv.org/abs/2405.02292}, 
}

@inproceedings{mobilealoha,
  author    = {Fu, Zipeng and Zhao, Tony Z. and Finn, Chelsea},
  title     = {Mobile ALOHA: Learning Bimanual Mobile Manipulation with Low-Cost Whole-Body Teleoperation},
  booktitle = {{Conference on Robot Learning (CoRL)}},
  year      = {2024},
}

@misc{aloha,
      title={Learning Fine-Grained Bimanual Manipulation with Low-Cost Hardware}, 
      author={Tony Z. Zhao and Vikash Kumar and Sergey Levine and Chelsea Finn},
      year={2023},
      eprint={2304.13705},
      archivePrefix={arXiv},
      primaryClass={cs.RO},
      url={https://arxiv.org/abs/2304.13705}, 
}

@inproceedings{robocasa2024,
  title={RoboCasa: Large-Scale Simulation of Everyday Tasks for Generalist Robots},
  author={Soroush Nasiriany and Abhiram Maddukuri and Lance Zhang and Adeet Parikh and Aaron Lo and Abhishek Joshi and Ajay Mandlekar and Yuke Zhu},
  booktitle={Robotics: Science and Systems (RSS)},
  year={2024}
}

@inproceedings{robocasa365,
  title={RoboCasa365: A Large-Scale Simulation Framework for Training and Benchmarking Generalist Robots},
  author={Soroush Nasiriany and Sepehr Nasiriany and Abhiram Maddukuri and Yuke Zhu},
  booktitle={International Conference on Learning Representations (ICLR)},
  year={2026}
}

@inproceedings{mimicgen,
    title={MimicGen: A Data Generation System for Scalable Robot Learning using Human Demonstrations},
    author={Mandlekar, Ajay and Nasiriany, Soroush and Wen, Bowen and Akinola, Iretiayo and Narang, Yashraj and Fan, Linxi and Zhu, Yuke and Fox, Dieter},
    booktitle={7th Annual Conference on Robot Learning},
    year={2023}
}

@inproceedings{dexmimicen,
      title     = {DexMimicGen: Automated Data Generation for Bimanual Dexterous Manipulation via Imitation Learning},
      author    = {Jiang, Zhenyu and Xie, Yuqi and Lin, Kevin and Xu, Zhenjia and Wan, Weikang and Mandlekar, Ajay and Fan, Linxi and Zhu, Yuke},
      booktitle = {2025 IEEE International Conference on Robotics and Automation (ICRA)},
      year      = {2025}
}

@misc{polaris,
      title={PolaRiS: Scalable Real-to-Sim Evaluations for Generalist Robot Policies}, 
      author={Arhan Jain and Mingtong Zhang and Kanav Arora and William Chen and Marcel Torne and Muhammad Zubair Irshad and Sergey Zakharov and Yue Wang and Sergey Levine and Chelsea Finn and Wei-Chiu Ma and Dhruv Shah and Abhishek Gupta and Karl Pertsch},
      year={2025},
      eprint={2512.16881},
      archivePrefix={arXiv},
      primaryClass={cs.RO},
      url={https://arxiv.org/abs/2512.16881}, 
}

@misc{r2r2r,
        title={Real2Render2Real: Scaling Robot Data Without Dynamics Simulation or Robot Hardware},
        author={Justin Yu and Letian Fu and Huang Huang and Karim El-Refai and Rares Andrei Ambrus and Richard Cheng and Muhammad Zubair Irshad and Ken Goldberg},
        year={2025},
        eprint={2505.09601},
        archivePrefix={arXiv},
        primaryClass={cs.RO},
        url={https://arxiv.org/abs/2505.09601},
  }

@misc{EmbodiedSplat,
      title={EmbodiedSplat: Personalized Real-to-Sim-to-Real Navigation with Gaussian Splats from a Mobile Device}, 
      author={Gunjan Chhablani and Xiaomeng Ye and Muhammad Zubair Irshad and Zsolt Kira},
      year={2025},
      eprint={2509.17430},
      archivePrefix={arXiv},
      primaryClass={cs.CV},
      url={https://arxiv.org/abs/2509.17430}, 
}

@inproceedings{re3sim,
  title={RE$^3$SIM: Generating High-Fidelity Simulation Data via 3D-Photorealistic Real-to-Sim for Robotic Manipulation},
  author={Han, Xiaoshen and Yu, Junqiu and Liu, Minghuan and Chen, Yilun and Lyu, Xiaoyang and Tian, Yang and Wang, Bolun and Zhang, Weinan and Zhang, Weinan and Pang, Jiangmiao},
  booktitle={IEEE International Conference on Robotics and Automation (ICRA)},
  year={2026}
}

@misc{planing,
      title={PLANING: A Loosely Coupled Triangle-Gaussian Framework for Streaming 3D Reconstruction}, 
      author={Changjian Jiang and Kerui Ren and Xudong Li and Kaiwen Song and Linning Xu and Tao Lu and Junting Dong and Yu Zhang and Bo Dai and Mulin Yu},
      year={2026},
      eprint={2601.22046},
      archivePrefix={arXiv},
      primaryClass={cs.CV},
      url={https://arxiv.org/abs/2601.22046}, 
}

@Article{3dgs,
      author       = {Kerbl, Bernhard and Kopanas, Georgios and Leimk{\"u}hler, Thomas and Drettakis, George},
      title        = {3D Gaussian Splatting for Real-Time Radiance Field Rendering},
      journal      = {ACM Transactions on Graphics},
      number       = {4},
      volume       = {42},
      month        = {July},
      year         = {2023},
      url          = {https://repo-sam.inria.fr/fungraph/3d-gaussian-splatting/}
}

@article{gsplat,
  title={gsplat: An open-source library for Gaussian splatting},
  author={Ye, Vickie and Li, Ruilong and Kerr, Justin and Turkulainen, Matias and Yi, Brent and Pan, Zhuoyang and Seiskari, Otto and Ye, Jianbo and Hu, Jeffrey and Tancik, Matthew and Angjoo Kanazawa},
  journal={Journal of Machine Learning Research},
  volume={26},
  number={34},
  pages={1--17},
  year={2025}
}

@article{trellis,
    title   = {Structured 3D Latents for Scalable and Versatile 3D Generation},
    author  = {Xiang, Jianfeng and Lv, Zelong and Xu, Sicheng and Deng, Yu and Wang, Ruicheng and Zhang, Bowen and Chen, Dong and Tong, Xin and Yang, Jiaolong},
    journal = {arXiv preprint arXiv:2412.01506},
    year    = {2024}
}

@inproceedings{nerf,
  title={NeRF: Representing Scenes as Neural Radiance Fields for View Synthesis},
  author={Ben Mildenhall and Pratul P. Srinivasan and Matthew Tancik and Jonathan T. Barron and Ravi Ramamoorthi and Ren Ng},
  year={2020},
  booktitle={ECCV},
}

@InProceedings{deepsdf,
author = {Park, Jeong Joon and Florence, Peter and Straub, Julian and Newcombe, Richard and Lovegrove, Steven},
title = {DeepSDF: Learning Continuous Signed Distance Functions for Shape Representation},
booktitle = {The IEEE Conference on Computer Vision and Pattern Recognition (CVPR)},
month = {June},
year = {2019}
}

@inproceedings{2DGS,
    title={2D Gaussian Splatting for Geometrically Accurate Radiance Fields},
    author={Huang, Binbin and Yu, Zehao and Chen, Anpei and Geiger, Andreas and Gao, Shenghua},
    publisher = {Association for Computing Machinery},
    booktitle = {SIGGRAPH 2024 Conference Papers},
    year      = {2024},
    doi       = {10.1145/3641519.3657428}
}

@InProceedings{4DGS,
    author    = {Wu, Guanjun and Yi, Taoran and Fang, Jiemin and Xie, Lingxi and Zhang, Xiaopeng and Wei, Wei and Liu, Wenyu and Tian, Qi and Wang, Xinggang},
    title     = {4D Gaussian Splatting for Real-Time Dynamic Scene Rendering},
    booktitle = {Proceedings of the IEEE/CVF Conference on Computer Vision and Pattern Recognition (CVPR)},
    month     = {June},
    year      = {2024},
    pages     = {20310-20320}
}

@misc{gssurvey,
      title={A Survey on 3D Gaussian Splatting}, 
      author={Guikun Chen and Wenguan Wang},
      year={2025},
      eprint={2401.03890},
      archivePrefix={arXiv},
      primaryClass={cs.CV},
      url={https://arxiv.org/abs/2401.03890}, 
}

@article{xsim,
    title={X-Sim: Cross-Embodiment Learning via Real-to-Sim-to-Real}, 
    author={Prithwish Dan and Kushal Kedia and Angela Chao and Edward Weiyi Duan and Maximus Adrian Pace and Wei-Chiu Ma and Sanjiban Choudhury},
    year={2025},
    eprint={2505.07096},
    archivePrefix={arXiv},
    primaryClass={cs.RO},
    url={https://arxiv.org/abs/2505.07096}, 
    }

@misc{splatsim,
            title={SplatSim: Zero-Shot Sim2Real Transfer of RGB Manipulation Policies Using Gaussian Splatting}, 
            author={Mohammad Nomaan Qureshi and Sparsh Garg and Francisco Yandun and David Held and George Kantor and Abhishesh Silwal},
            year={2024},
            eprint={2409.10161},
            archivePrefix={arXiv},
            primaryClass={cs.RO},
            url={https://arxiv.org/abs/2409.10161}, 
      }

@misc{realissim,
      title={Real-is-Sim: Bridging the Sim-to-Real Gap with a Dynamic Digital Twin}, 
      author={Jad Abou-Chakra and Lingfeng Sun and Krishan Rana and Brandon May and Karl Schmeckpeper and Niko Suenderhauf and Maria Vittoria Minniti and Laura Herlant},
      year={2025},
      eprint={2504.03597},
      archivePrefix={arXiv},
      primaryClass={cs.RO},
      url={https://arxiv.org/abs/2504.03597}, 
}

@article{twinaligner,
  author    = {Hongwei Fan and Hang Dai and Jiyao Zhang and Jinzhou Li and Qiyang Yan and Yujie Zhao and Mingju Gao and Jinghang Wu and Hao Tang and Hao Dong},
  title     = {TwinAligner: Visual-Dynamic Alignment Empowers Physics-aware Real2Sim2Real for Robotic Manipulation},
  year={2025},
  eprint={2512.19390},
  archivePrefix={arXiv},
  primaryClass={cs.RO},
  url={https://arxiv.org/abs/2512.19390},
}

@misc{DRAWER,
      title={DRAWER: Digital Reconstruction and Articulation With Environment Realism}, 
      author={Hongchi Xia and Entong Su and Marius Memmel and Arhan Jain and Raymond Yu and Numfor Mbiziwo-Tiapo and Ali Farhadi and Abhishek Gupta and Shenlong Wang and Wei-Chiu Ma},
      year={2025},
      eprint={2504.15278},
      archivePrefix={arXiv},
      primaryClass={cs.CV},
      url={https://arxiv.org/abs/2504.15278}, 
}

@inproceedings{
lehome,
title={LeHome: A Simulation Environment for Deformable Object Manipulation in Household Scenarios},
author={Zeyi Li and Jade Yang and Jingkai Xu and Shangbin Xie and Yuran Wang and Zhenhao Shen and Tianxing Chen and Yan Shen and Wenjun Li and Yukun Zheng and Chaorui Zhang and Ming Chen and Chen Xie and Ruihai Wu},
booktitle={IROS 2025 - 5th Workshop on RObotic MAnipulation of Deformable Objects: holistic approaches and challenges forward},
year={2025},
url={https://openreview.net/forum?id=rEDd1HorJl}
}

@article{3dgrt,
    author = {Nicolas Moenne-Loccoz and Ashkan Mirzaei and Or Perel and Riccardo de Lutio and Janick Martinez Esturo and Gavriel State and Sanja Fidler and Nicholas Sharp and Zan Gojcic},
    title = {3D Gaussian Ray Tracing: Fast Tracing of Particle Scenes},
    journal = {ACM Transactions on Graphics and SIGGRAPH Asia},
    year = {2024},
}

@article{3dgut,
    title={3DGUT: Enabling Distorted Cameras and Secondary Rays in Gaussian Splatting},
    author={Wu, Qi and Martinez Esturo, Janick and Mirzaei, Ashkan and Moenne-Loccoz, Nicolas and Gojcic, Zan},
    journal = {Conference on Computer Vision and Pattern Recognition (CVPR)},
    year={2025}
}

@misc{isaacsim,
  author       = {{NVIDIA Corporation}},
  title        = {{Isaac Sim}},
  year         = {2024},
  note         = {Version 5.1.0},
  howpublished = {\url{https://github.com/isaac-sim/IsaacSim}}
}

@article{isaaclab,
  title   = {Isaac Lab: A GPU-Accelerated Simulation Framework for Multi-Modal Robot Learning},
  author  = {Mittal, M. and Roth, P. and Tigue, J. and Richard, A. and Zhang, O. and Du, P. and Serrano-Mu{\~n}oz, A. and Yao, X. and Zurbr{\"u}gg, R. and Rudin, N. and Wawrzyniak, L. and Rakhsha, M. and Denzler, A. and Heiden, E. and Borovicka, A. and Ahmed, O. and Akinola, I. and Anwar, A. and Carlson, M. T. and Feng, J. Y. and Garg, A. and Gasoto, R. and Gulich, L. and Guo, Y. and Gussert, M. and Hansen, A. and Kulkarni, M. and Li, C. and Liu, W. and Makoviychuk, V. and Malczyk, G. and Mazhar, H. and Moghani, M. and Murali, A. and Noseworthy, M. and Poddubny, A. and Ratliff, N. and Rehberg, W. and Schwarke, C. and Singh, R. and Smith, J. L. and Tang, B. and Thaker, R. and Trepte, M. and Van Wyk, K. and Yu, F. and Millane, A. and Ramasamy, V. and Steiner, R. and Subramanian, S. and Volk, C. and Chen, C. and Jawale, N. and Kuruttukulam, A. V. and Lin, M. A. and Mandlekar, A. and Patzwaldt, K. and Welsh, J. and Lafleche, J. and Mo{\"e}nne-Loccoz, N. and Park, S. and Stepinski, R. and Van Gelder, D. and Amevor, C. and Carius, J. and Chang, J. and Chen, A. H. and Ciechomski, P. D. H. and Daviet, G. and Mohajerani, M. and von Muralt, J. and Reutskyy, V. and Sauter, M. and Schirm, S. and Shi, E. L. and Terdiman, P. and Vilella, K. and Widmer, T. and Yeoman, G. and Chen, T. and Grizan, S. and Li, C. and Li, L. and Smith, C. and Wiltz, R. and Alexis, K. and Chang, Y. and Fan, L. and Farshidian, F. and Handa, A. and Huang, S. and Hutter, M. and Narang, Y. and Pouya, S. and Sheng, S. and Zhu, Y. and Macklin, M. and Moravanszky, A. and Reist, P. and Guo, Y. and Hoeller, D. and State, G.},
  journal = {arXiv preprint arXiv:2511.04831},
  year    = {2025},
  doi     = {10.48550/arXiv.2511.04831}
}

@misc{polycam,
  author       = {{Polycam Inc.}},
  title        = {{Polycam}},
  year         = {2024},
  howpublished = {\url{https://poly.cam}},
  note         = {3D scanning application}
}

@article{sam,
  title={Segment Anything},
  author={Kirillov, Alexander and Mintun, Eric and Ravi, Nikhila and Mao, Hanzi and Rolland, Chloe and Gustafson, Laura and Xiao, Tete and Whitehead, Spencer and Berg, Alexander C. and Lo, Wan-Yen and Doll{\'a}r, Piotr and Girshick, Ross},
  journal={arXiv:2304.02643},
  year={2023}
}

@article{sam2,
  title={SAM 2: Segment Anything in Images and Videos},
  author={Ravi, Nikhila and Gabeur, Valentin and Hu, Yuan-Ting and Hu, Ronghang and Ryali, Chaitanya and Ma, Tengyu and Khedr, Haitham and R{\"a}dle, Roman and Rolland, Chloe and Gustafson, Laura and Mintun, Eric and Pan, Junting and Alwala, Kalyan Vasudev and Carion, Nicolas and Wu, Chao-Yuan and Girshick, Ross and Doll{\'a}r, Piotr and Feichtenhofer, Christoph},
  journal={arXiv preprint arXiv:2408.00714},
  url={https://arxiv.org/abs/2408.00714},
  year={2024}
}

@misc{sam3,
      title={SAM 3: Segment Anything with Concepts},
      author={Nicolas Carion and Laura Gustafson and Yuan-Ting Hu and Shoubhik Debnath and Ronghang Hu and Didac Suris and Chaitanya Ryali and Kalyan Vasudev Alwala and Haitham Khedr and Andrew Huang and Jie Lei and Tengyu Ma and Baishan Guo and Arpit Kalla and Markus Marks and Joseph Greer and Meng Wang and Peize Sun and Roman Rädle and Triantafyllos Afouras and Effrosyni Mavroudi and Katherine Xu and Tsung-Han Wu and Yu Zhou and Liliane Momeni and Rishi Hazra and Shuangrui Ding and Sagar Vaze and Francois Porcher and Feng Li and Siyuan Li and Aishwarya Kamath and Ho Kei Cheng and Piotr Dollár and Nikhila Ravi and Kate Saenko and Pengchuan Zhang and Christoph Feichtenhofer},
      year={2025},
      eprint={2511.16719},
      archivePrefix={arXiv},
      primaryClass={cs.CV},
      url={https://arxiv.org/abs/2511.16719},
}

@article{mobilesam,
  title={Faster Segment Anything: Towards Lightweight SAM for Mobile Applications},
  author={Zhang, Chaoning and Han, Dongshen and Qiao, Yu and Kim, Jung Uk and Bae, Sung-Ho and Lee, Seungkyu and Hong, Choong Seon},
  journal={arXiv preprint arXiv:2306.14289},
  year={2023}
}

@inproceedings{deva,
  title={Tracking Anything with Decoupled Video Segmentation},
  author={Cheng, Ho Kei and Oh, Seoung Wug and Price, Brian and Schwing, Alexander and Lee, Joon-Young},
  booktitle={ICCV},
  year={2023}
}

@article{groundingdino,
  title={Grounding dino: Marrying dino with grounded pre-training for open-set object detection},
  author={Liu, Shilong and Zeng, Zhaoyang and Ren, Tianhe and Li, Feng and Zhang, Hao and Yang, Jie and Li, Chunyuan and Yang, Jianwei and Su, Hang and Zhu, Jun and others},
  journal={arXiv preprint arXiv:2303.05499},
  year={2023}
}

@misc{GroundedSAM,
      title={Grounded SAM: Assembling Open-World Models for Diverse Visual Tasks}, 
      author={Tianhe Ren and Shilong Liu and Ailing Zeng and Jing Lin and Kunchang Li and He Cao and Jiayu Chen and Xinyu Huang and Yukang Chen and Feng Yan and Zhaoyang Zeng and Hao Zhang and Feng Li and Jie Yang and Hongyang Li and Qing Jiang and Lei Zhang},
      year={2024},
      eprint={2401.14159},
      archivePrefix={arXiv},
      primaryClass={cs.CV}
}

@article{EmbodiedSAM, 
      title={EmbodiedSAM: Online Segment Any 3D Thing in Real Time}, 
      author={Xiuwei Xu and Huangxing Chen and Linqing Zhao and Ziwei Wang and Jie Zhou and Jiwen Lu},
      journal={arXiv preprint arXiv:2408.11811},
      year={2024}
}

@inproceedings{autoseg3d,
  title={Online Segment Any 3D Thing as Instance Tracking},
  author={Hanshi Wang and Zijian Cai and Jin Gao and Yiwei Zhang and Weiming Hu and Ke Wang and Zhipeng Zhang},
  booktitle={The Thirty-ninth Annual Conference on Neural Information Processing Systems}
}

@article{GaussianGrouping,   title={Gaussian Grouping: Segment and Edit Anything in 3D Scenes},  author={Ye, Mingqiao and Danelljan, Martin and Yu, Fisher and Ke, Lei},  year={2023},  month={Dec},  language={en-US}  }

@article{saga,  
 title={Segment Any 3D Gaussians}, 
 author={Cen, Jiazhong and Fang, Jiemin and Yang, Chen and Xie, Lingxi and Zhang, Xiaopeng and Shen, Wei and Tian, Qi}, 
 year={2023}, 
 month={Dec}, 
 language={en-US} 
 }

@article{gaga,  
 title={Gaga: Group Any Gaussians via 3D-aware Memory Bank}, 
 author={Lyu, Weijie and Li, Xueting and Kundu, Abhijit and Tsai, Yi-Hsuan and Yang, Ming-Hsuan}, 
 year={2025}, 
 month={Mar}, 
 language={en-US} 
 }

@article{LangSplat,  
 title={LangSplat: 3D Language Gaussian Splatting}, 
 author={Qin, Minghan and Li, Wanhua and Zhou, Jiawei and Wang, Haoqian and Pfister, Hanspeter}, 
 year={2023}, 
 month={Dec}, 
 language={en-US} 
 }

@inproceedings{fastlgs,
 author = {Ji, Yuzhou and Zhu, He and Tang, Junshu and Liu, Wuyi and Zhang, Zhizhong and Tan, Xin and Xie, Yuan},
 title = {FastLGS: Speeding up Language Embedded Gaussians with Feature Grid Mapping},
 booktitle = {Proceedings of the AAAI Conference on Artificial Intelligence},
 year = {2025},
}

@inproceedings{objectgs,
  title={ObjectGS: Object-aware Scene Reconstruction and Scene Understanding via Gaussian Splatting},
  author={Zhu, Ruijie and Yu, Mulin and Xu, Linning and Jiang, Lihan and Li, Yixuan and Zhang, Tianzhu and Pang, Jiangmiao and Dai, Bo},
  booktitle={International Conference on Computer Vision (ICCV), 2025},
  year={2025}
}

@inproceedings{3dgic,
  title={3d gaussian inpainting with depth-guided cross-view consistency},
  author={Huang, Sheng-Yu and Chou, Zi-Ting and Wang, Yu-Chiang Frank},
  booktitle={Proceedings of the Computer Vision and Pattern Recognition Conference},
  pages={26704--26713},
  year={2025}
}

@article{lama,
  title={Resolution-robust Large Mask Inpainting with Fourier Convolutions},
  author={Suvorov, Roman and Logacheva, Elizaveta and Mashikhin, Anton and Remizova, Anastasia and Ashukha, Arsenii and Silvestrov, Aleksei and Kong, Naejin and Goka, Harshith and Park, Kiwoong and Lempitsky, Victor},
  journal={arXiv preprint arXiv:2109.07161},
  year={2021}
}

@article{inpaintanything,
  title={Inpaint Anything: Segment Anything Meets Image Inpainting},
  author={Yu, Tao and Feng, Runseng and Feng, Ruoyu and Liu, Jinming and Jin, Xin and Zeng, Wenjun and Chen, Zhibo},
  journal={arXiv preprint arXiv:2304.06790},
  year={2023}
}

@misc{lamarefine,
      title={Feature Refinement to Improve High Resolution Image Inpainting}, 
      author={Prakhar Kulshreshtha and Brian Pugh and Salma Jiddi},
      year={2022},
      eprint={2206.13644},
      archivePrefix={arXiv},
      primaryClass={cs.CV},
      url={https://arxiv.org/abs/2206.13644}, 
}

@article{diffusion,  
 title={Denoising Diffusion Probabilistic Models}, 
 author={Ho, Jonathan and Jain, Ajay and Abbeel, Pieter and Berkeley, UC}, 
 language={en-US} 
 }

@article{DeepUnsupervised,  
 title={Deep Unsupervised Learning using Nonequilibrium Thermodynamics}, 
 journal={arXiv: Learning,arXiv: Learning}, 
 author={Sohl-Dickstein, Jascha and Weiss, EricA. and Maheswaranathan, Niru and Ganguli, Surya}, 
 year={2015}, 
 month={Mar}, 
 language={en-US} 
 }

@inproceedings{openai,
  title={Language Models are Unsupervised Multitask Learners},
  author={Alec Radford and Jeff Wu and Rewon Child and David Luan and Dario Amodei and Ilya Sutskever},
  year={2019},
  url={https://api.semanticscholar.org/CorpusID:160025533}
}

@inproceedings{diffpoints,  
 title={Diffusion Probabilistic Models for 3D Point Cloud Generation}, 
 url={http://dx.doi.org/10.1109/cvpr46437.2021.00286}, 
 DOI={10.1109/cvpr46437.2021.00286}, 
 booktitle={2021 IEEE/CVF Conference on Computer Vision and Pattern Recognition (CVPR)}, 
 author={Luo, Shitong and Hu, Wei}, 
 year={2021}, 
 month={Jun}, 
 language={en-US} 
 }

@article{pointE,  
 title={Point-E: A System for Generating 3D Point Clouds from Complex Prompts}, 
 author={Nichol, Alex and Jun, Heewoo and Dhariwal, Prafulla and Mishkin, Pamela and Chen, Mark}, 
 year={2022}, 
 month={Dec}, 
 language={en-US} 
 }

@article{diffvolex,  
 title={Neural Wavelet-domain Diffusion for 3D Shape Generation}, 
 author={Hui, Ka-Hei and Li, Ruihui and Hu, Jingyu and Fu, Chi-Wing}, 
 year={2022}, 
 month={Sep}, 
 language={en-US} 
 }

@article{diffrf,  
 title={DiffRF: Rendering-Guided 3D Radiance Field Diffusion}, 
 journal={Cornell University - arXiv,Cornell University - arXiv}, 
 author={Muller, Norman and Siddiqui, Yawar and Porzi, Lorenzo and Bulò, SamuelRota and Kontschieder, Peter and Nießner, Matthias}, 
 year={2022}, 
 month={Dec}, 
 language={en-US} 
 }

@article{VolumeDiffusion,  
 title={VolumeDiffusion: Flexible Text-to-3D Generation with Efficient Volumetric Encoder}, 
 author={Tang, Zhicong and Gu, Shuyang and Wang, Chunyu and Zhang, Ting and Bao, Jianmin and Chen, Dong and Guo, Baining}, 
 year={2024}, 
 month={Apr}, 
 language={en-US} 
 }

@article{diffnerf,   title={Single-Stage Diffusion NeRF: A Unified Approach to 3D Generation and Reconstruction},  author={Chen, Hansheng and Gu, Jiatao and Chen, Anpei and Tian, Wei and Tu, Zhuowen and Liu, Lingjie and Su, Hao},  year={2023},  month={Apr},  language={en-US}  }

@article{difftri,   title={3D Neural Field Generation using Triplane Diffusion},  author={Shue, J.Ryan and Chan, EricRyan and Po, Ryan and Ankner, Zachary and Wu, Jiajun and Wetzstein, Gordon},  year={2022},  month={Nov},  language={en-US}  }

@article{Rodin,   title={Rodin: A Generative Model for Sculpting 3D Digital Avatars Using Diffusion},  author={Wang, Tengfei and Zhang, Bo and Zhang, Ting and Gu, Shuyang and Bao, Jianmin and Baltrusaitis, Tadas and Shen, Jingjing and Chen, Dong and Wen, Fang and Chen, Qifeng and Guo, Baining and Research, Microsoft},  language={en-US}  }

@article{RodinHD,   title={RodinHD: High-Fidelity 3D Avatar Generation with Diffusion Models},  author={Zhang, Bowen and Cheng, Yiji and Wang, Chunyu and Zhang, Ting and Yang, Jiaolong and Tang, Yansong and Zhao, Feng and Chen, Dong and Guo, Baining},  year={2024},  month={Jul},  language={en-US}  }

@article{GVGEN,  
 title={GVGEN: Text-to-3D Generation with Volumetric Representation}, 
 author={He, Xianglong and Chen, Junyi and Peng, Sida and Huang, Di and Li, Yangguang and Huang, Xiaoshui and Yuan, Chun and Ouyang, Wanli and He, Tong}, 
 language={en-US} 
 }

@article{gaussiancube,
  title={GaussianCube: Structuring Gaussian Splatting using Optimal Transport for 3D Generative Modeling},
  author={Zhang, Bowen and Cheng, Yiji and Yang, Jiaolong and Wang, Chunyu and Zhao, Feng and Tang, Yansong and Chen, Dong and Guo, Baining},
  journal={arXiv preprint arXiv:2403.19655},
  year={2024}
}

@article{polygen,   title={PolyGen: An Autoregressive Generative Model of 3D Meshes},  journal={Cornell University - arXiv,Cornell University - arXiv},  author={Nash, Charlie and Ganin, Yaroslav and Eslami, S.M.Ali and Battaglia, PeterW.},  year={2020},  month={Feb},  language={en-US}  }

@misc{meshanything,
  title={MeshAnything: Artist-Created Mesh Generation with Autoregressive Transformers},
  author={Yiwen Chen and Tong He and Di Huang and Weicai Ye and Sijin Chen and Jiaxiang Tang and Xin Chen and Zhongang Cai and Lei Yang and Gang Yu and Guosheng Lin and Chi Zhang},
  year={2024},
  eprint={2406.10163},
  archivePrefix={arXiv},
  primaryClass={cs.CV}
}

@article{sam3d,
      title={SAM 3D: 3Dfy Anything in Images}, 
      author={SAM 3D Team and Xingyu Chen and Fu-Jen Chu and Pierre Gleize and Kevin J Liang and Alexander Sax and Hao Tang and Weiyao Wang and Michelle Guo and Thibaut Hardin and Xiang Li and Aohan Lin and Jiawei Liu and Ziqi Ma and Anushka Sagar and Bowen Song and Xiaodong Wang and Jianing Yang and Bowen Zhang and Piotr Dollár and Georgia Gkioxari and Matt Feiszli and Jitendra Malik},
      year={2025},
      eprint={2511.16624},
      archivePrefix={arXiv},
      primaryClass={cs.CV},
      url={https://arxiv.org/abs/2511.16624}, 
}

@article{omnipart,
        title={Omnipart: Part-aware 3d generation with semantic decoupling and structural cohesion},
        author={Yang, Yunhan and Zhou, Yufan and Guo, Yuan-Chen and Zou, Zi-Xin and Huang, Yukun and Liu, Ying-Tian and Xu, Hao and Liang, Ding and Cao, Yan-Pei and Liu, Xihui},
        journal={arXiv preprint arXiv:2507.06165},
        year={2025}
}

@article{physxanything,
  title={PhysX-Anything: Simulation-Ready Physical 3D Assets from Single Image},
  author={Cao, Ziang and Hong, Fangzhou and Chen, Zhaoxi and Pan, Liang and Liu, Ziwei},
  journal={arXiv preprint arXiv:2511.13648},
  year={2025}
}

@article{physx3d,
  title={PhysX-3D: Physical-Grounded 3D Asset Generation},
  author={Cao, Ziang and Chen, Zhaoxi and Pan, Liang and Liu, Ziwei},
  journal={arXiv preprint arXiv:2507.12465},
  year={2025}
}

@article{ScreenedPoisson,
  title={Screened poisson surface reconstruction},
  author={ Kazhdan, Michael  and  Hoppe, Hugues },
  journal={Acm Transactions on Graphics},
  volume={32},
  number={3},
  pages={1-13},
  year={2013},
}

@article{Poisson,
  title={Poisson surface reconstruction},
  author={ Kazhdan, M },
  year={2006},
}

@misc{embodiedgen,
      title={EmbodiedGen: Towards a Generative 3D World Engine for Embodied Intelligence},
      author={Xinjie Wang and Liu Liu and Yu Cao and Ruiqi Wu and Wenkang Qin and Dehui Wang and Wei Sui and Zhizhong Su},
      year={2025},
      eprint={2506.10600},
      archivePrefix={arXiv},
      primaryClass={cs.RO},
      url={https://arxiv.org/abs/2506.10600},
}

@misc{uninavid,
        title={Uni-NaVid: A Video-based Vision-Language-Action Model for Unifying Embodied Navigation Tasks}, 
        author={Jiazhao Zhang and Kunyu Wang and Shaoan Wang and Minghan Li and Haoran Liu and Songlin Wei and Zhongyuan Wang and Zhizheng Zhang and He Wang},
        year={2024},
        journal = {arXiv preprint arXiv:2412.06224}
      }

@article{scenemaker,
        title={SceneMaker: Open-set 3D Scene Generation with Decoupled De-occlusion and Pose Estimation Model},
        author={Shi, Yukai and Li, Weiyu and Wang, Zihao and Li, Hongyang and Chen, Xingyu and Tan, Ping and Zhang, Lei},
        journal={arXiv preprint arXiv:2512.10957},
        year={2025}
      }

@misc{FlexWorld,
      title={FlexWorld: Progressively Expanding 3D Scenes for Flexiable-View Synthesis}, 
      author={Luxi Chen and Zihan Zhou and Min Zhao and Yikai Wang and Ge Zhang and Wenhao Huang and Hao Sun and Ji-Rong Wen and Chongxuan Li},
      year={2025},
      eprint={2503.13265},
      archivePrefix={arXiv},
      primaryClass={cs.CV},
      url={https://arxiv.org/abs/2503.13265}, 
}

@inproceedings{gscream,
     title={GScream: Learning 3D Geometry and Feature Consistent Gaussian Splatting for Object Removal},
     author={Wang, Yuxin and Wu, Qianyi and Zhang, Guofeng and Xu, Dan},
     booktitle={ECCV},
     year={2024}
     }

@article{barron2022mipnerf360,
    title={Mip-NeRF 360: Unbounded Anti-Aliased Neural Radiance Fields},
    author={Jonathan T. Barron and Ben Mildenhall and 
            Dor Verbin and Pratul P. Srinivasan and Peter Hedman},
    journal={CVPR},
    year={2022}
}

@inproceedings{schoeps2017cvpr,
  author = {Thomas Sch\"ops and Johannes L. Sch\"onberger and Silvano Galliani and Torsten Sattler and Konrad Schindler and Marc Pollefeys and Andreas Geiger},
  title = {A Multi-View Stereo Benchmark with High-Resolution Images and Multi-Camera Videos},
  booktitle = {Conference on Computer Vision and Pattern Recognition (CVPR)},
  year = {2017}
}

@misc{insta360x5,
  title        = {Insta360 X5},
  author       = {{Insta360}},
  year         = {2025},
  howpublished = {\url{https://www.insta360.com/}},
  note         = {Accessed: 2026-05-09}
}

@misc{postshot,
  title        = {Postshot},
  author       = {{Jawset Visual Computing}},
  year         = {2025},
  howpublished = {\url{https://www.jawset.com/}},
  note         = {Accessed: 2026-05-09}
}

@misc{piper_arm,
  title        = {Piper Robotic Arm},
  author       = {{AgiBot}},
  year         = {2025},
  howpublished = {\url{https://www.agibot.com/}},
  note         = {Accessed: 2026-05-09}
}
\clearpage
\appendix
\section*{\centering\bfseries APPENDIX}
% \addcontentsline{toc}{section}{APPENDIX}

% \section{Candidate Image Ranking}
% \label{appendix a}
% For user-selected point set $P$, let $\Pi_f$ be visible projected points on image $f$. We compute
% \begin{equation}
% c_f=\frac{|\Pi_f|}{|P|},\qquad
% A_f=\left(\max_{q\in\Pi_f}u_q-\min_{q\in\Pi_f}u_q\right)
% \left(\max_{q\in\Pi_f}v_q-\min_{q\in\Pi_f}v_q\right),
% \end{equation}
% \begin{equation}
% s_f=A_f\cdot c_f,
% \end{equation}
% and rank candidate images by $s_f$.

\begin{figure*}[h]
    \centering
    \includegraphics[width=1.05\linewidth, trim={5cm 7cm 2.5cm 0.5cm}, clip]{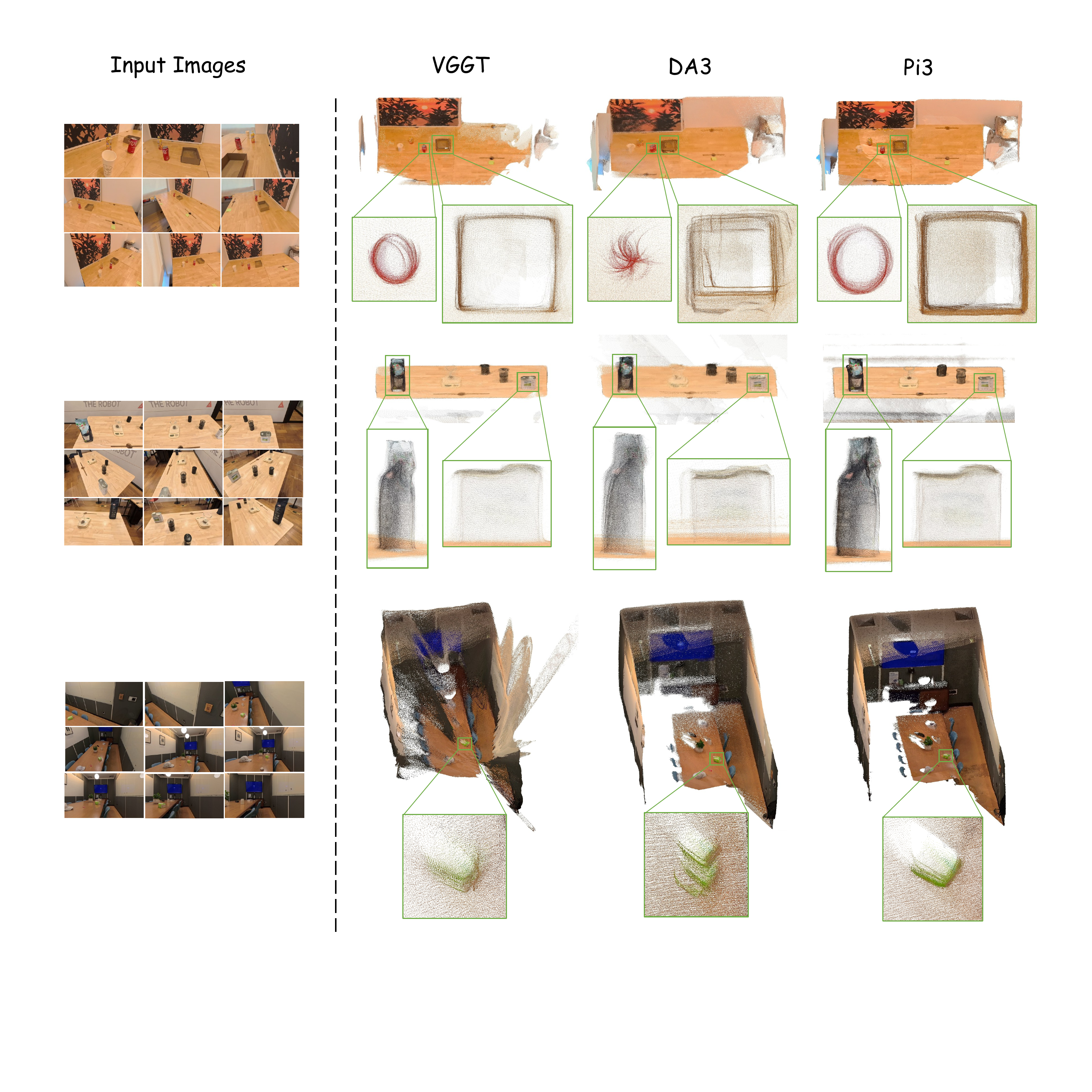}
    \caption{Qualitative comparison of different initialization methods. Zoom-in patches highlight the texture blurriness and layering artifacts present in feed-forward predictions (VGGT, DA3, Pi3) compared to the input reference.}
    \label{fig:qualitative_results}
\end{figure*}

\section{3DGS training}
\label{appendix b}
After mask-guided point-cloud cleaning, we feed inpainted scene images and COLMAP data to gsplat for 3D Gaussian reconstruction. The model representation, training objective, and depth rendering are defined as follows.

\subsection{Training Details}
The reconstruction loss used in gsplat is
\begin{equation}
\mathcal{L}_{\mathrm{gs}}
=
(1-\lambda_{\mathrm{ssim}})
\left\|\hat{\mathbf{I}}-\mathbf{I}\right\|_1
+
\lambda_{\mathrm{ssim}}
\left(1-\mathrm{SSIM}(\hat{\mathbf{I}},\mathbf{I})\right),
\end{equation}
where $\mathbf{I}$ is the ground-truth inpainted image, $\hat{\mathbf{I}}$ is the rendered image, and $\lambda_{\mathrm{ssim}}$ is the SSIM weight.

\subsection{Depth Rendering for Mesh Extraction}
After training, we render expected depth (ED) for each camera view:
\begin{equation}
\hat{D}(\mathbf{r})
=
\frac{\sum_{k} w_k(\mathbf{r})\, z_k(\mathbf{r})}
{\sum_{k} w_k(\mathbf{r})},
\end{equation}
where $\mathbf{r}$ is a camera ray, $w_k(\mathbf{r})$ is the contribution weight of Gaussian $k$ on that ray, and $z_k(\mathbf{r})$ is its corresponding depth.  
The resulting $\hat{D}$ is saved as raw depth maps and used for subsequent TSDF/Poisson mesh extraction.

\section{Weighting Scheme}
\label{appendix c}
For each candidate object crop $i$ with a width $w_i$ and height $h_i$, we compute a weighted score
\begin{equation}
S_i
=
0.20\,\bar{A}_i
+0.15\,F_i
+0.10\,C_i
+0.20\,E_i
+0.10\,R_i
+0.25\,Q_i ,
\end{equation}
and rank candidates in descending order of $S_i$.

The six terms are defined as follows.

\paragraph{(1) Normalized object area}
\begin{equation}
\bar{A}_i=\frac{A_i}{\max_j A_j}
\end{equation}
,where $A_i$ means the number of foreground pixels in candidate $i$.

\paragraph{(2) Fill ratio}
\begin{equation}
F_i=\frac{A_i}{B_i},\qquad
B_i=w_i h_i,
\end{equation}
where $w_i,h_i$ are the width and height of the object bounding box.

\paragraph{(3) Center prior}
\begin{equation}
C_i=\frac{1}{1+\frac{d_i}{L_i}},\qquad
L_i=\max(W_i,H_i),
\end{equation}
where $d_i$ is the Euclidean distance between the object center and image center, and $W_i,H_i$ are image width and height.

\paragraph{(4) Boundary-cut penalty}
\begin{equation}
E_i=
\begin{cases}
1.0, & \text{object does not touch image boundary},\\
0.5, & \text{object touches boundary (potential truncation)}.
\end{cases}
\end{equation}

\paragraph{(5) Aspect-ratio regularity}
\begin{equation}
R_i=\frac{1}{1+\left|\ln\left(\frac{w_i}{h_i}\right)\right|}.
\end{equation}

\paragraph{(6) Effective-resolution score}
\begin{equation}
Q_i=\min\!\left(1,\frac{\mathrm{ER}_i}{S_{\mathrm{out}}}\right),
\end{equation}
where $\mathrm{ER}_i=\max(w_i,h_i)$ is effective object resolution and $S_{\mathrm{out}}$ is the target TRELLIS input size (e.g., 512).

\paragraph{Selection with viewpoint diversity}
After scoring, candidates are greedily selected with a minimum frame-gap constraint to avoid near-duplicate views. If the selected set is smaller than the target count, the frame-gap constraint is gradually relaxed to fill the quota.

\section{Limitation Details}
\label{app:limitation}
Due to limitations in certain underlying technologies, our system still has aspects that can be further improved. We discuss these from two perspectives: feed-forward 3D reconstruction and semantic segmentation within 3DGS.

\paragraph{Semantic Segmentation within 3DGS} Existing system do not support real-time object segmentation after reconstruction. If users wish to extract new objects from an already reconstructed scene, they must rerun the entire workflow, which is time-consuming. If the issue of unclear object boundaries in semantic segmentation within 3DGS, mentioned in \ref{Introduction} and \ref{segmentation}, can be solved, it would become possible to build a practical system that supports real-time segmentation.

\paragraph{Feed-forward 3D Reconstruction}
Although feed-forward networks offer a highly efficient and lightweight alternative to traditional SfM pipelines (e.g., COLMAP) for 3DGS initialization, they are fundamentally bottlenecked by their capacity to reconstruct high-frequency geometric details. Compared to the sparse yet globally consistent point clouds generated by COLMAP, feed-forward predictions occasionally exhibit discrete approximations, manifesting as layering artifacts and quantization errors. Because the optimization landscape of 3D Gaussian Splatting is highly sensitive to initialization for escaping local minima, these coarse geometric priors can result in over-smoothed textures and a degradation of fine-grained structural fidelity in the final novel views. Comprehensive quantitative evaluations and qualitative visualizations validating these limitations, including experiments evaluated on several representative feed-forward methods, are provided in the supplementary material. Future research will focus on bridging this precision gap---such as through the integration of lightweight refinement modules or stricter multi-view photometric constraints---while preserving inherent efficiency.

\section{Additional Experiments}
\label{appendix d}

\subsection{Analysis of Initialization Precision}

As discussed in the App.\ref{app:limitation}, while feed-forward initialization significantly accelerates the pipeline, it currently trails traditional SfM methods like COLMAP in capturing high-frequency geometric features and fine-grained textures. To empirically demonstrate this limitation and ensure reproducibility, we conduct comprehensive evaluations using representative feed-forward networks (VGGT, DA3, and Pi3) on a single NVIDIA A800 GPU.

\paragraph{Experimental Setup.} We evaluate the methods on the Mip-NeRF 360~\cite{barron2022mipnerf360} and ETH3D~\cite{schoeps2017cvpr} datasets. Following standard evaluation protocols, images in the Mip-NeRF 360 dataset are downsampled by a factor of 4 for outdoor scenes and 2 for indoor scenes, whereas the ETH3D dataset is evaluated at full resolution. To assess rendering quality, we employ standard novel view synthesis metrics (PSNR, SSIM, and LPIPS). Furthermore, to accurately reflect the efficiency of the initialization and rendering pipeline, our reported FPS indicates the validation rendering speed---calculated as the inverse of the average rasterization time per image---strictly distinguishing it from training or video playback FPS.

\paragraph{Results and Discussion.} Table~\ref{tab:metrics_comparison} presents our quantitative analysis, clearly illustrating the trade-off between efficiency and rendering fidelity. While feed-forward models achieve superior validation rendering speeds, they lag behind COLMAP in reconstruction precision. This precision gap is particularly evident in the LPIPS metric, which is highly sensitive to high-frequency structural loss, quantitatively validating our observation that the coarse priors from current feed-forward networks lead to over-smoothed final renderings. Fundamentally, this performance bottleneck arises because feed-forward architectures rely on generalized training priors and single-pass regression. They inherently lack the rigorous, scene-specific multi-view photometric consistency enforced by iterative bundle adjustment in optimization-based SfM methods. Figure~\ref{fig:qualitative_results} further provides qualitative zoom-in comparisons to directly illustrate the geometric limitations of these feed-forward representations. When compared to the reference input images, the predictions from VGGT, DA3, and Pi3 consistently exhibit noticeable texture blurriness, loss of fine details, and discrete structural approximations (e.g., layering artifacts) in complex local regions. Consequently, to guarantee the highest quality novel view synthesis and provide a mathematically robust geometric prior that prevents 3D Gaussian Splatting from converging to sub-optimal local minima, we ultimately employ COLMAP as the initialization module for our pipeline.

\begin{table*}[htbp]
\centering
\caption{Quantitative comparison of different methods on Mip-NeRF 360 and ETH3D datasets.}
\label{tab:metrics_comparison}
{
\begin{tabular}{l cccc cccc}
\toprule
\multirow{2}{*}{Method} & \multicolumn{4}{c}{Mip-NeRF 360} & \multicolumn{4}{c}{ETH3D} \\
\cmidrule(lr){2-5} \cmidrule(lr){6-9}
 & PSNR $\uparrow$ & SSIM $\uparrow$ & LPIPS $\downarrow$ & FPS $\uparrow$ & PSNR $\uparrow$ & SSIM $\uparrow$ & LPIPS $\downarrow$ & FPS $\uparrow$ \\
\midrule
VGGT   & 21.8846 & 0.5954 & 0.3322 & 35.7612 & 16.4594 & 0.7031 & 0.5537 & 6.9671 \\
DA3    & 20.8818 & 0.5446 & 0.3334 & 45.7707 & 17.0423 & 0.6903 & 0.5109 & 9.2058 \\
Pi3    & 21.8899 & 0.5570 & 0.2787 & 42.3389 & 17.5147 & 0.6901 & 0.4838 & 8.7631 \\
COLMAP & 29.2393 & 0.8878 & 0.1238 & 32.2384 & 19.8355 & 0.8118 & 0.4128 & 6.5092 \\
\bottomrule
\end{tabular}
} 
\end{table*}

\section{Dataset Creation}
\label{appendix e}
We scanned the scene using 1 to 3 Insta360 X5\cite{insta360x5} panoramic cameras. Each resulting panoramic video was then split into 2 to 5 viewpoints. We then stitched them together in the order of the high, middle, and low panoramic cameras to obtain the scene image, and fed it into Postshot\cite{postshot} to recover the camera poses and sparse point cloud.

\end{document}